\documentclass{CVM}

\usepackage{color, colortbl}
\usepackage{float}
\definecolor{lightgray}{gray}{0.9}
\usepackage{xcolor}
\usepackage{bm}
\usepackage{amsmath}
\usepackage{multirow}
\usepackage{makecell}
\usepackage{arydshln}
\usepackage{pifont}
\usepackage{soul}
\usepackage{bbm}
\usepackage{amsfonts}
\usepackage{amssymb}
\usepackage{gensymb}

\newcommand{\peng}[1]{{#1}}

\usepackage{tabularx}

\CVMsetup{
type      = {Research/Review Article},
doi       = {CVM.XXXX},
title     = {Gaussian-plus-SDF SLAM: High-fidelity 3D Reconstruction at 150+ fps},
author    = {Zhexi Peng$^{1,2}$, Kun Zhou\cor{}$^{1,2}$, and Tianjia Shao$^{1,2}$\\
},
runauthor = {Zhexi Peng, Kun Zhou, Tianjia Shao},
abstract  = {
While recent Gaussian-based SLAM methods achieve photorealistic reconstruction from RGB-D data, their computational performance remains a critical bottleneck. State-of-the-art techniques operate at less than 20 fps, significantly lagging behind geometry-based approaches like KinectFusion (hundreds of fps). This limitation stems from the heavy computational burden: modeling scenes requires numerous Gaussians and complex iterative optimization to fit RGB-D data; insufficient Gaussian counts or optimization iterations cause severe quality degradation. To address this, we propose a Gaussian-SDF hybrid representation, combining a colorized signed distance field (SDF) for smooth geometry and appearance with 3D Gaussians to capture underrepresented details. The SDF is efficiently constructed via RGB-D fusion (as in geometry-based methods), while Gaussians undergo iterative optimization. 
Our representation enables significant Gaussian reduction ($50\%$ fewer) by avoiding full-scene Gaussian modeling, and efficient Gaussian optimization ($75\%$ fewer iterations) through targeted appearance refinement.
Building upon this representation, we develop GPS-SLAM (Gaussian-plus-SDF SLAM), a real-time 3D reconstruction system achieving over 150 fps on real-world Azure Kinect sequences, faster by an order-of-magnitude than state-of-the-art techniques while maintaining comparable reconstruction quality. 
The source code and data are available at https://gapszju.github.io/GPS-SLAM.
},
keywords  = {3D reconstruction, Gaussian splatting, SLAM, RGBD cameras, 3D scanning, signed distance fields},
copyright = {The Author(s)},
}

\begin{document}

\maketitle

\enlargethispage{-3pt}
\begin{figure}[b] \vskip -4mm
\small\renewcommand\arraystretch{1.3}
\begin{tabular}{p{80.5mm}} \toprule\\ \end{tabular}
\vskip -4.5mm \noindent \setlength{\tabcolsep}{1pt}
\begin{tabular}{p{3.5mm}p{80mm}}
$1\quad $ & State Key Lab of CAD\&CG, Zhejiang University, Hangzhou, 310058, China. E-mails: \{zhexipeng, kunzhou, tjshao\} @zju.edu.cn.\\
$2\quad $ & Hangzhou Research Institute of AI and Holographic Technology, Hangzhou, 310015, China.\\
&\hspace{-5mm} Manuscript received: 2025-07-31; accepted: 2025-09-16
\end{tabular}
\end{figure}

\section{Introduction}\label{sec:intro}
Recent advances in Gaussian-based RGB-D SLAM have demonstrated impressive capabilities in jointly reconstructing high-quality geometry and photorealistic appearance\peng{~\cite{splatam, GauS-SLAM, peng2024rtgslam}}. However, achieving real-time performance remains a significant challenge. The core limitation stems from the computationally expensive online optimization of 3D Gaussians required by these methods. To accurately fit observations based on image loss, a large number of Gaussians must be optimized, a process typically demanding many iterations for convergence. For instance, RTG-SLAM~\cite{peng2024rtgslam}, the state-of-the-art Gaussian-based RGB-D SLAM system, achieves real-time reconstruction at 17 fps (59 ms per frame), with Gaussian optimization alone consuming an average of 38 ms per frame. While reducing the number of optimization iterations improves speed, it inevitably compromises reconstruction quality. GS-ICP SLAM~\cite{GS-ICP_SLAM}, for example, reports real-time speeds (e.g., 100 fps) by decoupling camera tracking and scene mapping into separate threads, where mapping operates slower than tracking. Consequently, reconstruction terminates prematurely upon tracking completion, before optimization converges, leading to significantly degraded reconstruction quality, as we report later in Table~\ref{tab:time_and_memory_and_render} and Fig.~\ref{fig:Indoor_compare}.

Conversely, traditional geometry-based RGB-D SLAM methods---exemplified by KinectFusion~\cite{kinectfusion} and its derivatives~\cite{slam-octree,vox_hashing,bundlefusion, InfiniTAM_ISMAR_2015}---achieve real-time performance at hundreds of fps. These approaches represent scenes using volumetric truncated signed distance fields (SDF) and fuse each incoming depth and RGB frame directly into the SDF volume. The fusion process is highly efficient, averaging approximately 0.1 ms per frame. However, it is well known that this direct fusion strategy introduces significant color artifacts, including blur and texture holes resulting from depth occlusion or sensor limitations. Consequently, the photometric quality of the reconstructions produced by these methods is generally inferior to that achieved by Gaussian-based approaches.

In this paper, we introduce a novel method that combines the ultra-fast reconstruction speed of geometry-based SLAM with the photorealistic rendering quality of Gaussian-based SLAM. Our core contribution is a hybrid scene representation, Gaussian-plus-SDF, comprising a core SDF representation providing base geometry and color, and an optimizable Gaussian overlay dedicated to appearance refinement. The SDF serves as the foundational scene model, delivering 3D structure and initial color. The 3D Gaussians are then optimized via photometric loss to correct color inaccuracies within the SDF (e.g., blur, missing data) and to model high-frequency appearance details beyond the SDF's capacity.
A key insight underpinning our approach is that the SDF already provides a geometrically coherent scene initialization. This fundamentally simplifies the role of the Gaussians: instead of modeling the entire scene geometry and appearance from scratch, their task is reduced to efficient appearance correction and enhancement. This shift yields two significant advantages: the number of required Gaussians is drastically reduced, and the optimization complexity is substantially lowered, enabling rapid convergence. Consequently, the computational overhead introduced by Gaussian optimization is minimized, while retaining the same efficient SDF fusion speed as KinectFusion~\cite{kinectfusion, InfiniTAM_ISMAR_2015}. This synergy enables our method to achieve ultra-fast, high-fidelity reconstruction.

Technically, Gaussian-plus-SDF is a hybrid representation combining a signed distance field (SDF) with a 3D Gaussian radiance field. The SDF volume stores truncated distance values and initial RGB colors on surface voxels, while a sparse set of 3D Gaussians distributed around the reconstructed surface forms a volumetric radiance field to correct residual color errors. The rendering pipeline operates in two stages. In the first stage,  standard ray-casting is performed on the SDF volume to generate a surface depth map and an initial SDF-rendered color image. In the second stage, 3D Gaussians are rendered with depth-culling using the surface depth map. Their colors are accumulated per pixel into the color image via order-independent blending. This sorting-free Gaussian rendering significantly speeds up the forward rendering process~\cite{ye2025gaussian}. More importantly, by shifting parallelization from the pixel level to Gaussian level, we avoid inefficient atomic operations during gradient accumulation in backpropagation, and achieve significant speedup in optimization.

We leveraging our Gaussian-SDF hybrid representation in GPS-SLAM (Gaussian-plus-SDF SLAM), an ultra-fast RGB-D SLAM system for real-time 3D reconstruction. Built upon the InfiniTAM framework~\cite{InfiniTAM_ISMAR_2015, InfiniTAM_ECCV_2016}, it processes each input frame through three stages: estimating camera pose via SDF-based alignment, integrating depth and RGB data into the SDF volume, and optimizing 3D Gaussians via photometric loss minimization, dynamically managed by our high-efficiency Gaussian insertion and removal strategy.
We have evaluated the system extensively on three benchmarks: Replica~\cite{replica19arxiv}, TUM-RGBD~\cite{TUM}, and ScanNet++~\cite{yeshwanthliu2023scannetpp}, plus diverse real-world scenes. Compared to state-of-the-art Gaussian-based RGB-D SLAM systems, our method achieves much faster reconstruction (150+ fps for real-world scenes and 250+ fps for Replica scenes), which is an order of magnitude faster than current best methods, while maintaining comparable high-quality reconstruction and photorealistic appearance.

\begin{figure}[t]
    \centering
\includegraphics[width=\columnwidth]{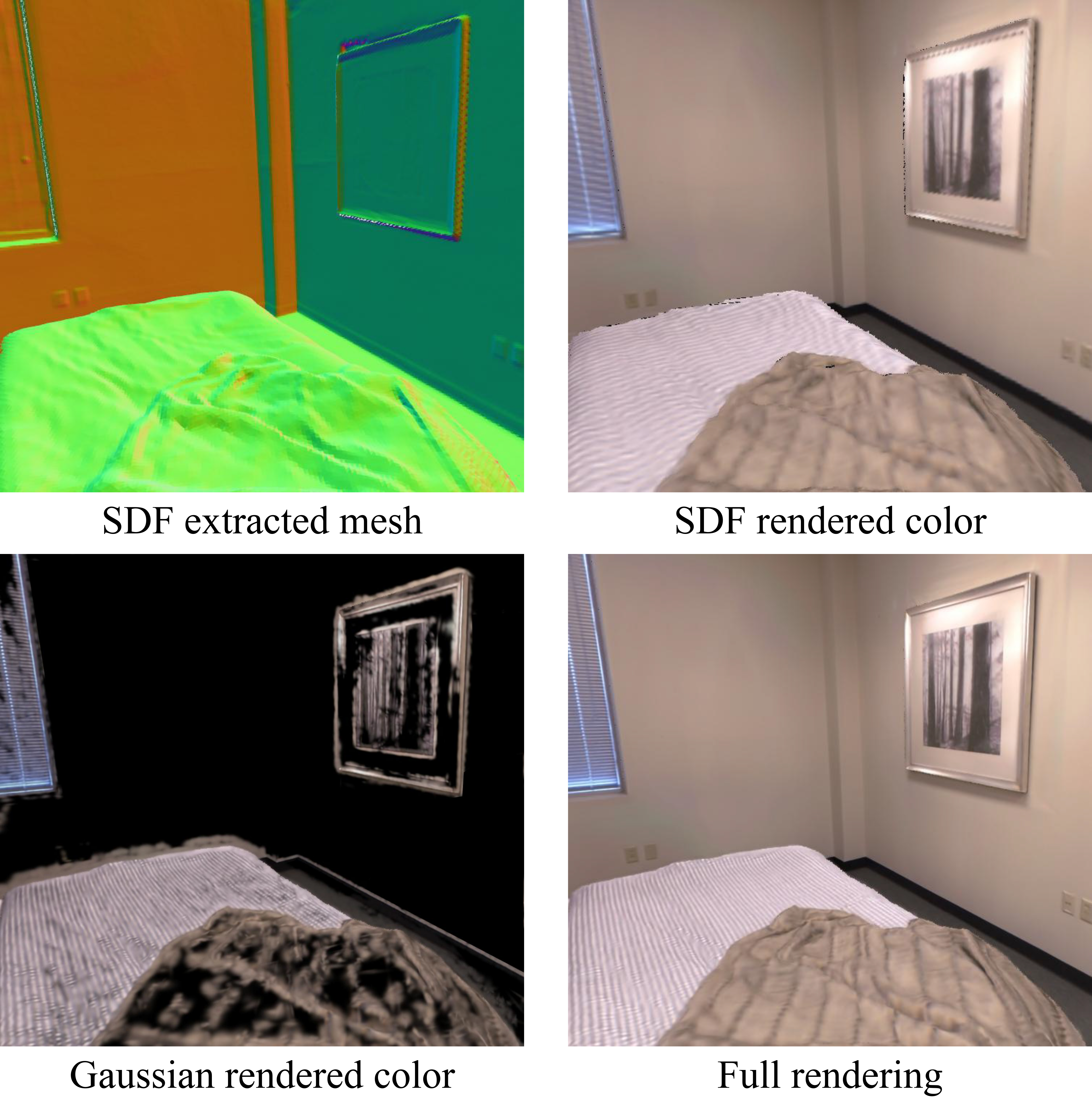}
    \caption{In our Gaussian-plus-SDF representation, the SDF provides 3D structure and initial color, while the 3D Gaussians are optimized to correct residual color errors. This allows us to achieve high-fidelity reconstruction with a small number of Gaussians, enabling ultra-fast scene reconstruction.}
    \label{fig:inllustration}
\end{figure}

\section{Related Work}\label{sec:related}
\subsection{Classical RGB-D volumetric SLAM}
Extensive research has been conducted on online 3D reconstruction with RGB-D cameras. Following the advent of KinectFusion~\cite{kinectfusion}, volumetric mapping utilizing SDFs has emerged as a popular reconstruction representation. The original volumetric mapping was limited to small scenes due to its large memory overhead. To address this issue, octree-based surface representations~\cite{ZENG2013126, 10.1145/2070781.2024182, 6907127} have been proposed as alternatives to uniform voxel grids to reduce memory cost and  computation time. On the other hand, \cite{vox_hashing} utilizes a spatial hash table to represent scenes. InfiniTAM redesigns the data structure to allow for much faster read and write operations, achieving hundreds of fps. However, these existing methods focus on geometry reconstruction but struggle with realistic texture due to simple color averaging in the voxels. Our method addresses this problem by extending InfiniTAM with radiance fields, achieving photorealistic rendering during ultra-fast reconstruction.

\subsection{NeRF-based RGB-D dense SLAM}
Following the breakthrough success of neural radiance fields (NeRFs)~\cite{NeRF20}, researchers have increasingly explored the integration of NeRFs into RGB-D dense SLAM systems. A pioneering effort in this direction is iMap~\cite{imap}, which introduced a NeRF SLAM method employing a single MLP for scene representation. Subsequent advances include NICE-SLAM~\cite{nice_slam}, which utilizes hierarchical feature grids and pre-trained MLPs for scene decoding. Vox-Fusion~\cite{vox_fusion} adopts a voxel-based neural implicit surface representation, efficiently stored using octrees. ESLAM~\cite{eslam} and Co-SLAM~\cite{co_slam} leverage multi-resolution feature grids and hash grids, respectively, for scene representation. Currently, the state-of-the-art NeRF SLAM system, HS-SLAM~\cite{Gong2025HS-SLAM}, uses a hybrid encoding network that combines hash grids, tri-planes, and one-blob to improve the completeness and smoothness of reconstruction. On the other hand, Point-SLAM~\cite{point_slam} and Loopy-SLAM~\cite{liso2024loopy} employ neural point embedding for high-quality dense reconstruction. Despite achieving high visual quality, these methods are limited by the computational burden of volume rendering, hindering real-time performance and increasing memory consumption.

\subsection{Gaussian-based RGB-D dense SLAM}
3D Gaussians~\cite{3DGS} achieve high-quality novel view synthesis and real-time rendering through differentiable rasterization. This has led to research into employing 3D Gaussians as a map representation within SLAM systems~\cite{splatam, gaussianslam, gs_dense_slam, hu2024cg}. RTG-SLAM proposes a compact Gaussian representation, enabling reconstruction speeds of around 15 fps for real scenes. However, this still falls short of the speeds of traditional SLAM methods. GS-ICP SLAM achieves a reconstruction speed of 100 fps by completely decoupling Gaussian optimization from camera tracking. However, it is not a purely real-time system because it does not optimize the newly captured frame. GauS-SLAM~\cite{GauS-SLAM} delivers superior tracking precision and rendering fidelity based on 2D Gaussian surfels, but the reconstruction speed is quite slow. GSFusion~\cite{GSFusion} leverages  octree-based SDF to reconstruct geometry, while Gaussians are used solely to represent texture. This approach significantly reduces the number of Gaussians required, but the complex Gaussian initialization mechanism limits its runtime performance and results in poor rendering quality. 

\subsection{Other work}
Our work is also related to the offline 3D reconstruction approach GSDF~\cite{GSDF}, which introduces a dual-branch architecture combining neural SDF and 3DGS to enhance reconstruction and rendering. While their SDF and 3DGS in GSDF are two independent representations, our Gaussian-plus-SDF is a hybrid representation in which the SDF represents the 3D scene with smooth \peng{geometry} and initial colors, and the 3D Gaussians correct color errors. 
Our representation is inspired by the most recent work GESs (Gaussian-enhanced surfels)~\cite{ye2025gaussian}, which employs a bi-scale approach for radiance field rendering, using surfels for coarse scene representation and 3D Gaussians to enhance fine-scale details, achieving  3.6$\times$ the rendering speed of 3DGS. However, GESs is also designed for offline reconstruction and is difficult to apply to online reconstruction. In contrast, our representation can be optimized very quickly and enables real-time online reconstruction.

\section{Method}\label{sec:method}

\begin{figure*}[t!]
    \centering
    \includegraphics[width= 0.95\textwidth]{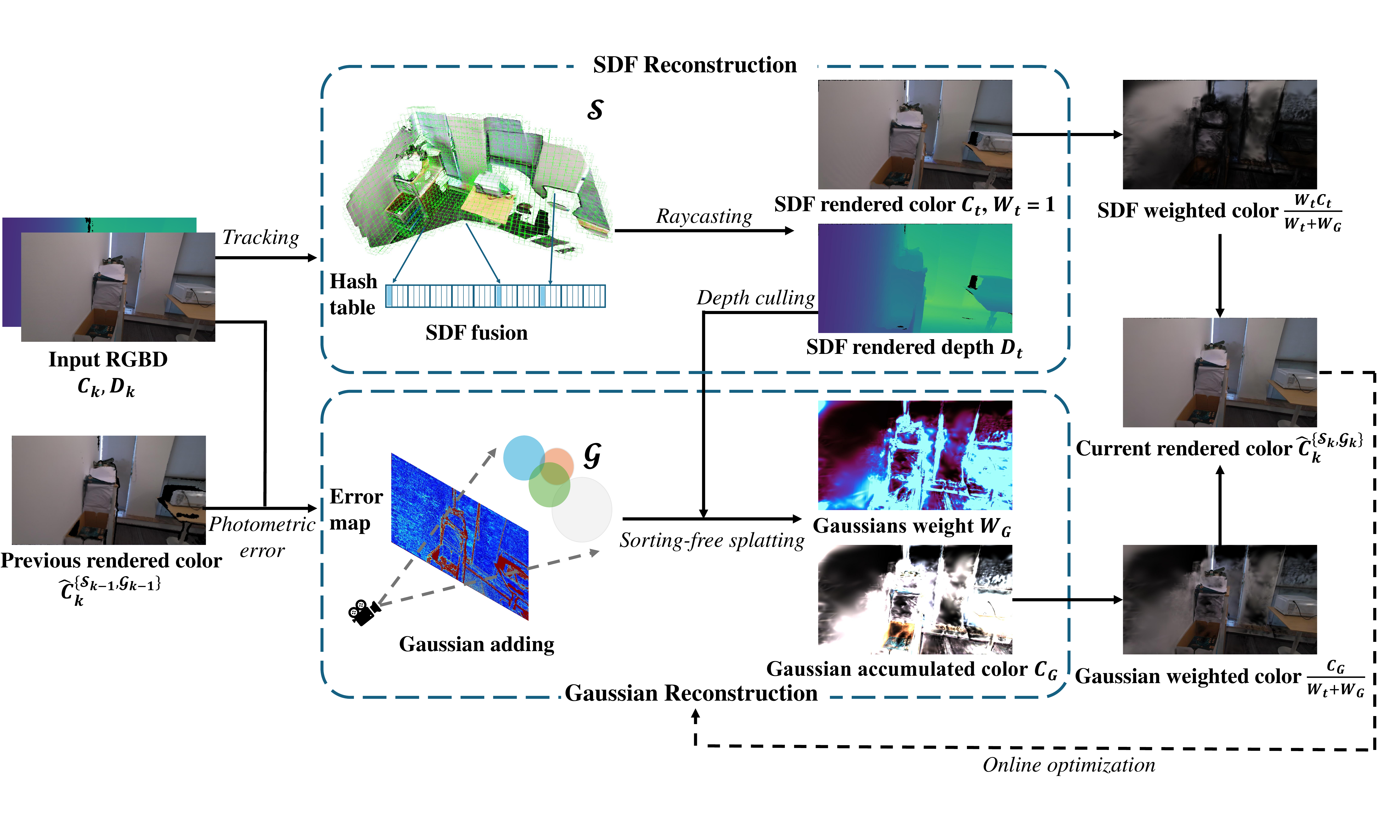}
    \caption{Overview of our GPS-SLAM system. Given an RGB-D image as input, standard SDF fusion is performed to update the SDF and color values in a global hash table. Then, we sample pixels within regions exhibiting significant color errors and add Gaussians to these locations. Online optimization for 3D Gaussians is executed based on our Gaussian-plus-SDF rendering.}
    \label{fig:pipeline}
\end{figure*}

\begin{figure*}[!t]
    \centering
\includegraphics[width=\textwidth]{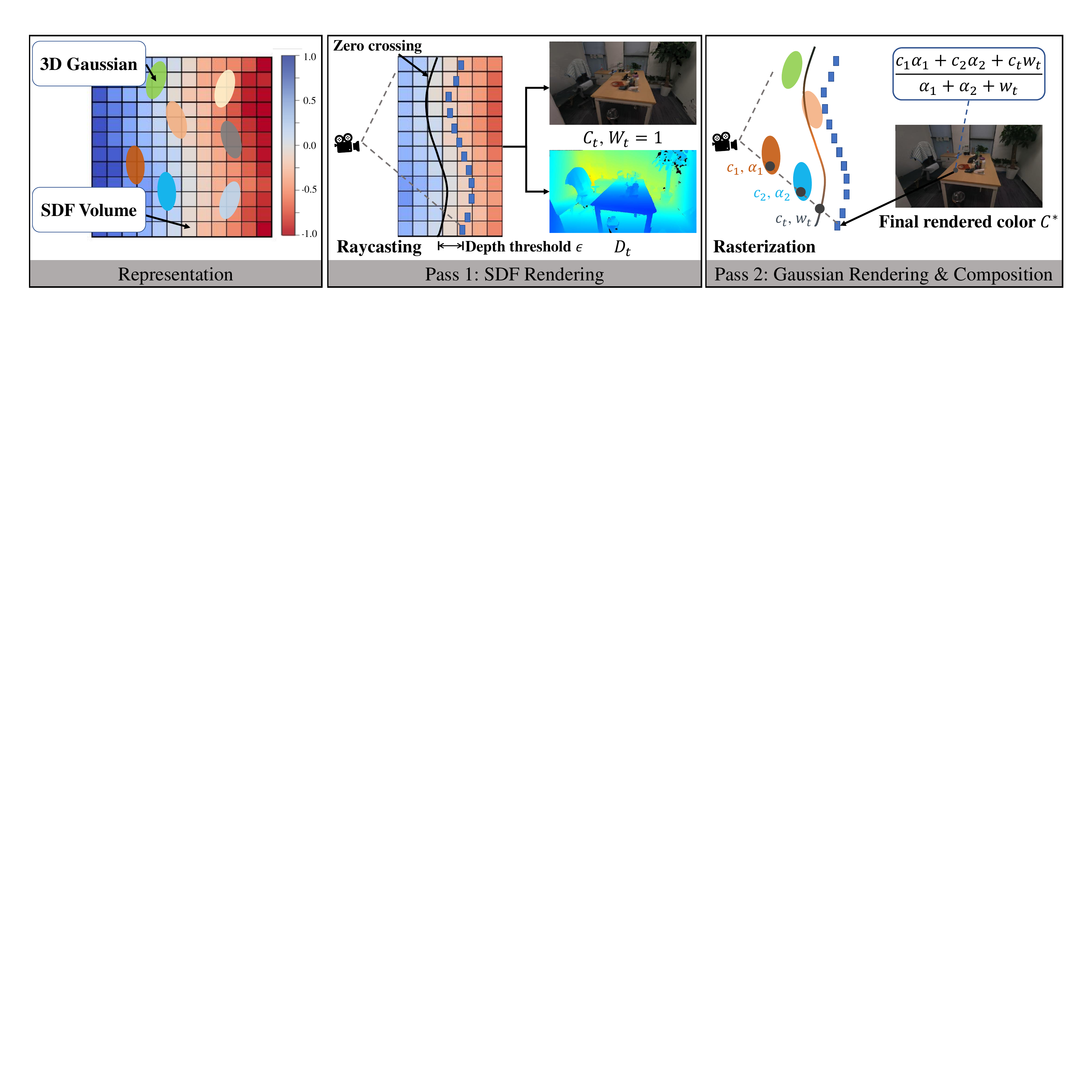}
    \caption{Gaussian-plus-SDF representation and rendering process. Our representation consists of an SDF volume $\mathcal{S}$ and a set of 3D Gaussians $\mathcal{G}=\{\mathbf{p}_i, \sigma_i, \mathbf{r}_i, \mathbf{s}_i, \mathbf{SH}_i\}_{i=1}^M$, with  properties: position $\mathbf{p}_i$, maximum opacity $\sigma_i$, scaling $\mathbf{s}_i$, rotation $\mathbf{r}_i$ and SH coefficients $\mathbf{SH}_i$. The rendering process consists of two passes. Firstly,  standard per-pixel ray-casting is performed on the SDF volume to obtain a color map $C_t$ and depth map $D_t$. Secondly, we splat the Gaussians to the screen, and their colors and weights are blended together independently of order with depth testing based on the SDF-rendered depth map $D_t$. A small positive threshold $\epsilon$ is used to prevent incorrect truncation. The accumulated Gaussian color and SDF-rendered color are combined as a weighted average to get the final image.}
    \label{fig:repre_render}
\end{figure*}

An overview of our SLAM pipeline is provided in Fig.~\ref{fig:pipeline}. In Sec.~\ref{Map_Representation}, we first introduce our Gaussian-plus-SDF representation, and the corresponding rendering process (see Fig.~\ref{fig:repre_render}). Next,  in Sec.~\ref{SLAM_process}, we describe in detail the entire online reconstruction process based on the Gaussian-plus-SDF representation.

\subsection{Gaussian-plus-SDF}\label{Map_Representation}
Gaussian-plus-SDF is a hybrid representation of the signed distance field and the radiance field. It uses an SDF volume to store SDF values and RGB colors on voxels, representing the 3D scene surfaces with initial colors. A set of 3D Gaussians surrounding the surface forms a volumetric radiance field to correct residual color errors.

Following KinectFusion, the SDF volume $\mathcal{S}$ is defined as a discretization of the SDF with a specified resolution of voxels, where each voxel $\mathbf{p}$ stores its truncated signed distance to the scene surface $d(\mathbf{p})$ and its RGB color $\mathbf{c}(\mathbf{p})$.   
The 3D Gaussians $\mathcal{G}$ are defined as $\mathcal{G}=\{\mathbf{p}_i, \sigma_i, \mathbf{r}_i, \mathbf{s}_i, \mathbf{SH}_i\}_{i=1}^M$, with  properties: position $\mathbf{p}_i$, maximum opacity $\sigma_i$, scaling $\mathbf{s}_i$, rotation $\mathbf{r}_i$ and \peng{spherical harmonic (SH)} coefficients $\mathbf{SH}_i$\peng{, following~\cite{3DGS}}.

The rendering of Gaussian-plus-SDF consists of two passes. In the first pass, 
standard per-pixel ray-casting is performed on the SDF volume~\cite{vox_hashing, InfiniTAM_ISMAR_2015}, where the ray for each pixel marches through the SDF voxels to find the zero crossing, as in~\cite{InfiniTAM_ISMAR_2015}. 
When the zero-crossing point for a pixel $\mathbf{u}$ is found, we obtain the 3D surface vertex $\mathbf{V}^{g}(\mathbf{u})$ in the world coordinate system. 
Afterwards, we compute the RGB value $\mathbf{C}_t(\mathbf{u})$ for each pixel by linearly interpolating the colors of the eight neighboring voxels of $\mathbf{V}^{g}(\mathbf{u})$. The depth value $D_t(\mathbf{u})$ of the pixel is also obtained by projecting $\mathbf{V}^{g}(\mathbf{u})$ onto the image plane. In this way, the SDF-rendered image $\mathbf{C}_t$ and the depth map $D_t$ are available after the first rendering pass. 

In the second pass, the Gaussians are splatted onto the screen, with their colors and weights accumulated for each pixel in an order-independent way. We conduct depth culling on the Gaussians during Gaussian accumulation, using the SDF-rendered depth map $D_t$. Gaussians whose centers are occluded by the scene surface are ignored during rendering.
Specifically, the accumulated Gaussian color and weight for a pixel $\hat{\mathbf{x}}$ are defined as
\begin{eqnarray}
    \mathbf{C}_G(\hat{\mathbf{x}})&=&\sum_{i=1}^K[\mathbbm{1}(d_i<D_t(\hat{\mathbf{x}})+\epsilon)]\mathbf{c}_i\alpha_i(\hat{\mathbf{x}}), 
\label{equ:accu_color}    
\\
    W_G(\hat{\mathbf{x}})&=&\sum_{i=1}^K[\mathbbm{1}(d_i<D_t(\hat{\mathbf{x}})+\epsilon)]\alpha_i(\hat{\mathbf{x}}),
\label{equ:accu_weight}  
\\
    \alpha_i(\hat{\mathbf{x}})&=&\sigma_i\exp\left(-\frac{(\hat{\mathbf{x}}-\hat{\mathbf{p}_i})^T\bm{\Sigma}_\mathrm{2D}^{-1}(\hat{\mathbf{x}}-\hat{\mathbf{p}_i})}{2}\right),
\label{equ:opacity}
\end{eqnarray}
where $D_t(\hat{\mathbf{x}})$ is the depth of pixel $\hat{\mathbf{x}}$ in the SDF-rendered depth map $D_t$, and $d_i$ is the depth of the Gaussian's center. $\mathbbm{1}(\cdot)$ is the indicator function, and $\epsilon$ is a small positive threshold introduced to prevent  3D Gaussians distributed close to the surface from being wrongly truncated. 
$\bm{\Sigma}_\mathrm{2D}$ is the covariance matrix of projected Gaussians in pixel space.  $\hat{\mathbf{p}_i}$ is the projected Gaussian center, and $\mathbf{c}_i$ is the Gaussian color in the viewing direction corresponding to the image. $\alpha_i(\hat{\mathbf{x}})$ is clamped to $0$ if $\alpha_i(\hat{\mathbf{x}})<1/255$, as in 3DGS~\cite{3DGS}. 

After the two-pass rendering, we obtain the final rendered image $\mathbf{C}^*$ as a weighted combination of the SDF-rendered image and the Gaussian rendered image
\begin{equation}
    \mathbf{C}^* = \frac{\mathbf{C}_tW_t+\mathbf{C}_G}{W_t+W_G}.
\label{equ:final_color}
\end{equation}
Here, $W_t = 1$ is the fixed weight for the SDF color. 

The Gaussian colors are accumulated and normalized by weights, so the rendering process of Gaussian-plus-SDF is entirely sorting-free, which also helps GPS-SLAM reach ultra-fast performance. First, in forward rendering, the computation bottleneck of Gaussian sorting in 3DGS is bypassed, boosting the rendering speed. More importantly, during backpropagation in the online optimization in SLAM, 3DGS requires the launching of threads per-pixel and traversing Gaussians in a back-to-front order to compute gradients. This results in slow gradient accumulation for each Gaussian because of the inefficiency of atomic addition. In contrast, we estimate the affected range for each Gaussian based on its pixel-space radius and launch threads per-Gaussian to directly accumulate gradients within each thread, significantly improving computational speed.
To address the varying sizes of different Gaussians, we implement a carefully designed thread scheduling mechanism to achieve optimal thread utilization. Further implementation details can be found in  Appendix~\ref{app:optimization_details}.

\subsection{Online Reconstruction Process}\label{SLAM_process}
As  Fig.~\ref{fig:pipeline} shows, our online reconstruction system has two parts. In the SDF reconstruction part, we perform camera tracking and update the signed distance field, acquiring the 3D structure and initial colors. In the Gaussian reconstruction part, we conduct online Gaussian insertion, Gaussian optimization and Gaussian removal, obtaining high-fidelity scene appearance with details.

\subsubsection{SDF reconstruction}
\paragraph{SDF fusion} Given the $k$-th frame of an RGB-D video stream (i.e., an RGB image $\mathbf{C}_k$ and a depth map $D_k$), we compute the local vertex map $\mathbf{V}_k^l$ and the local normal map $\mathbf{N}_k^l$ using the intrinsic camera parameters. Using the estimated camera pose $\mathbf{T}_{g,k}$, $\mathbf{V}_k^l$ and $\mathbf{N}_k^l$ are transformed into the global vertex map $\mathbf{V}_k^g$ and global normal map $\mathbf{N}_k^g$ in  world space. Following InfiniTAM, we perform standard SDF fusion to update the SDF and color values in a global hash table. Afterwards,  ray-casting is performed, yielding the ray-casting vertex map $\mathbf{V}_k^{*,g}$ and normal map $\mathbf{N}_k^{*,g}$ in  world space. Then the SDF-rendered color map $\mathbf{C}_{tk}$ is computed by interpolation using $\mathbf{V}_k^{*,g}$, and the SDF-rendered depth map $D_{tk}$ is computed by projecting $\mathbf{V}_k^{*,g}$ onto the image plane.

\paragraph{Camera tracking} We adopt the standard ICP method for camera tracking~\cite{kinectfusion, InfiniTAM_ISMAR_2015}, which minimizes the point-to-plane distance as
\begin{equation}
    E(\boldsymbol{\xi}) = \sum \left\| \big(\mathbf{T}_{g, k} \mathbf{V}_k^l(\mathbf{u}) -\mathbf{V}_{k-1}^{*,g}(\hat{\mathbf{u}})\big) \cdot \mathbf{N}_{k-1}^{*,g}(\hat{\mathbf{u}})\right\|.
\end{equation}
Here $\boldsymbol{\xi}$ is the Lie algebra representation of the estimated transformation $\mathbf{T}_{g, k}$. 
$\mathbf{u}$ is the pixel in the current frame, and $\hat{\mathbf{u}} = \pi(\mathbf{K} \mathbf{T}_{g,k-1}\mathbf{V}_k^l(\mathbf{u})$) is the projection of $\mathbf{V}_k^l(\mathbf{u})$ into the previous frame. $\pi(\cdot)$ performs perspective projection. A resolution hierarchy of the depth map is used in our implementation to improve  convergence behavior.

\subsubsection{Gaussian reconstruction}
Based on the SDF reconstruction, we perform Gaussian reconstruction to correct the residual color errors. Since adjacent views often contain a large amount of redundant information, we perform Gaussian reconstruction at certain time intervals $\delta_k$.  Gaussian reconstruction includes Gaussian insertion, Gaussian optimization, and Gaussian removal. Note that  Gaussian reconstruction does not affect SDF reconstruction, so the two reconstruction processes can be executed in parallel for speed.

\paragraph{Gaussian insertion} Instead of densifying Gaussians based on gradient magnitude \cite{3DGS}, we design an efficient Gaussian insertion strategy explicitly targeting regions with appearance details not well represented  by SDF.
Specifically, for the $k$-th frame with  SDF-rendered image $\mathbf{C}_{tk}$ and depth map $\mathbf{D}_{tk}$, we apply the second-pass rendering using the existing Gaussians in the scene to obtain the final rendered image $\mathbf{C}^*_k$ and the Gaussian weight map $W_{Gk}$.
Then a mask $M_{\widetilde{\mathbf{{u}}}}$ is created to determine for which pixels a Gaussian should be added
\begin{align}
    &M_{\widetilde{\mathbf{{u}}}} = \{\widetilde{\mathbf{u}}\big||\mathbf{C}^*_k(\mathbf{u})- \mathbf{C}_k(\mathbf{u})| > \delta_\mathbf{c} \,\text{and }\, W_{Gk} < \delta_W\}.
\end{align}
$M_{\widetilde{\mathbf{u}}}$ represents those pixels with apparent color errors but insufficient Gaussian counts. $\delta_\mathbf{c} = 0.05$ is the threshold of color error and $\delta_W = 4$ is the threshold of Gaussian weights.

\peng{While $M_{\widetilde{\mathbf{{u}}}}$ indicates regions where Gaussians might be needed, the substantial GPU memory requirement of inserting Gaussians for all $M_{\widetilde{\mathbf{{u}}}}$ significantly impedes real-time performance. Therefore, we uniformly sample 25\% pixels on $M_{\widetilde{\mathbf{{u}}}}$ and create a new Gaussian for each sampled pixel $\mathbf{u}$. For each Gaussian, the 3D position $\mathbf{p}$ and the zero-order component of $\mathbf{SH}$ are initialized to $\mathbf{V}_k^{*,g}(\mathbf{u})$ and $\mathbf{C}_k(\mathbf{u})$. The Gaussian's shape is initialized as a flat surface-aligned disc, as in RTG-SLAM. The shortest axis of the Gaussian is initialized to align with $\mathbf{N}_k^{*,g}(\mathbf{u})$ and the initial opacity is set to $0.5$. Please refer to  Appendix~\ref{app:initialization_details} for more details.}

\begin{figure}[t!]
    \centering
\includegraphics[width=\columnwidth]{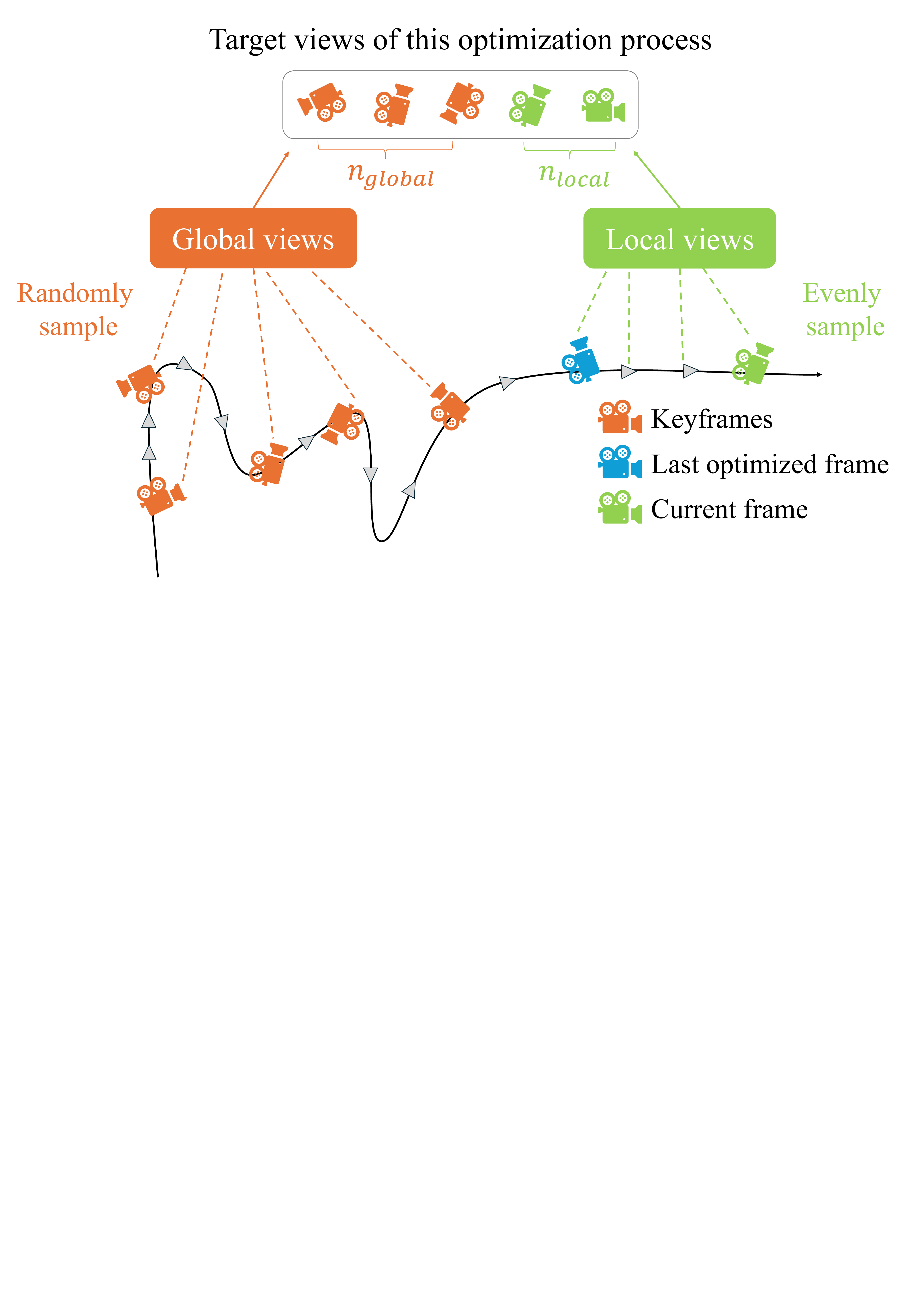}
    \caption{Optimization view selection. We randomly sample $n_\mathrm{global}$ views from global keyframes and evenly select $n_\mathrm{local}$ views from local frames during an optimization time interval $\delta_k$. This approach allows us to optimize  recent frames while mitigating both overfitting and catastrophic forgetting.}
    \label{fig:view_sample_method}
\end{figure}

\paragraph{Gaussian optimization}
After adding and initializing Gaussians in  world space, their optimization is guided by selected views. If only recent frames are optimized, this may result in over-fitting issues, degrading the quality of the global Gaussian map. On the other hand, since the SDF is continuously being fused and updated, some Gaussians that were optimized in the past may not fit the latest SDF. Therefore, we use keyframes for Gaussian optimization to ensure global consistency in optimization. Our keyframe selection strategy is inspired by \cite{Real-Time-High-Accuracy}. The keyframe list is constructed based on the camera motion. If the rotation angle relative to the last keyframe exceeds a threshold $\delta_\mathrm{angle}$, or the relative translation exceeds $\delta_\mathrm{move}$, we add a new keyframe. We randomly sample $n_\mathrm{global}$ global keyframes from the  keyframe history and evenly select $n_\mathrm{local}$ recent local frames from the frames during an optimization time interval $\delta_k$. These frames collectively form the target views for this optimization, as shown in Fig.~\ref{fig:view_sample_method}.

After determining the views to be optimized, we use the latest SDF to perform ray-casting on these views, recording the SDF-rendered color images and depth maps to avoid repeated ray-casting during the optimization process. We perform Gaussian optimization based on the color loss between the input and rendered RGB images. We use the $L_1$ loss for optimization
\begin{equation}
    L_\mathrm{color} = |\mathbf{C}^*_k - \mathbf{C}_k|.
\end{equation}

\paragraph{Gaussian removal} Although we  consider historical views in the optimization, the number of views used in each optimization is still small to ensure speed. This could potentially lead to overfitting, resulting in some excessively large Gaussians that can significantly impact  global quality. Meanwhile, the optimization process can also generate some Gaussians with too small scale or opacity. These Gaussians make relatively little contribution to the reconstruction and can be considered  redundant. After each Gaussian optimization, we delete Gaussians in the scene that satisfy the following criteria
\begin{equation}
\begin{split}
M_\mathcal{G}=\{&\mathcal{G}_i| \sigma_i < \delta_{\sigma},\text{~or~} \max(\mathbf{s}^{1,2,3}_i) > \delta_{\mathbf{s}_\mathrm{max}}, \\
&\text{or~}\max(\mathbf{s}^{1,2,3}_i) < \delta_{\mathbf{s}_\mathrm{min}}\},
\end{split}
\end{equation}
where $\delta_{\mathbf{s}_\mathrm{max}} = 0.1$ is the threshold for excessively large Gaussians and $\delta_{\mathbf{s}_\mathrm{min}} = 0.003$ is the threshold for too small Gaussians. $\delta_{\sigma} = 0.005$ is the threshold for low opacity Gaussians.

\section{Evaluation}\label{sec:result}
\subsection{Experimental Setup}
\subsubsection{Implementation details}
We implemented our SLAM system on a desktop computer with an AMD 9950X3D CPU and a 4GB Nvidia RTX 4090 GPU. We developed the entire system in C++ using  the Libtorch framework and wrote custom CUDA kernels for rasterization and backpropagation. In all experiments, Gaussian optimization was launched every 10 frames, with 20 iterations performed each time. The voxel size for SDF fusion was set to 0.5 cm. Please refer to Appendix~\ref{app:implement_details} for further details.

\subsubsection{Datasets} We evaluated our method and all baselines on three public datasets: Replica, TUM-RGBD, ScanNet++, and a our own indoor dataset captured by scanning. Replica is the simplest benchmark due to its synthetic, highly accurate, and complete RGB-D images. 
TUM-RGBD is a widely used dataset in the SLAM field for evaluating tracking accuracy because it provides accurate camera poses from an external motion capture system. We tested on three sequences (\emph{fr1\_desk}, \emph{fr2\_xyz}, \emph{fr3\_office}), which have been commonly used in previous studies.
ScanNet++ is a large-scale dataset that combines high-quality and commodity-level geometry and color capture of indoor scenes. Its depth maps are rendered from models reconstructed from laser scanning. Unlike other benchmarks, the camera poses in ScanNet++ are very far apart from one another. Following~\cite{splatam, GauS-SLAM}, two challenging sequences (b20a261fdf, 8b5caf3398) were sampled for evaluation. We also scanned five real-world indoor scenes with an Azure Kinect RGB-D camera to build our own indoor dataset.

\subsubsection{Baselines} We compared our method to existing state-of-the-art Gaussian RGB-D SLAM methods including GSFusion, GS-ICP SLAM, and RTG-SLAM. We also compared our method to the concurrent work GauS-SLAM. For GS-ICP SLAM, the number of map iterations varies depending on  tracking speed. The authors provided two cases: one with  tracking speed limited to 30 fps, and another with unlimited tracking speed. We used the unlimited mode to evaluate performance under the fastest conditions. We reproduced the results using the published code and conducted all experiments on the same desktop computer. 

\subsection{Evaluation of Online Reconstruction}
\subsubsection{Time/memory performance}
To ensure that our system can offer immediate feedback during real-time scanning, we compared its speed and GPU memory usage to those of all baseline approaches: see Table~\ref{tab:time_and_memory_and_render}. Across all evaluated datasets, our method consistently delivers ultra-fast and stable performance, significantly outperforming other methods, including GSFusion, which is also based on SDF fusion and implemented in C++. Our system runs at only 79 fps on ScanNet++ because of its much higher resolution (1752$\times$1168, which is 2.2$\times$ that of \peng{standard RGB-D images}). Please note that GS-ICP SLAM and GauS-SLAM exhaust the entire 24GB of GPU memory when processing the high-resolution ScanNet++ dataset. Furthermore, our method achieves 250 fps on the synthetic Replica Dataset. \peng{The Replica scenes are larger than our indoor scenes, yet the overall memory overhead is similar. It shows that the Gaussian reconstruction on the Replica Dataset consumes less memory than that on our indoor dataset.} This is attributed to the high-quality SDF reconstruction, which reduces the number of Gaussians required. Fig. \ref{fig::render_method_ablation} later illustrates the differences in SDF and Gaussian rendering results between real-world and synthetic \peng{datasets}.

\begin{table}[t!]
    \caption{Comparison of runtime performance and rendering quality on the Replica, ScanNet++ and our indoor datasets. Our indoor dataset was scanned by ourselves with an Azure Kinect RGB-D camera, and contains five real-world indoor scenes. Our system runs at only 79 fps on ScanNet++ because of its much higher resolution (1752$\times$1168, which is 2.2$\times$ that of \peng{standard RGB-D images}). The numbers for memory indicate usage in MB; \ding{53} means out of memory.}
    \label{tab:time_and_memory_and_render}
    \centering
\setlength\tabcolsep{0pt}
\begin{tabular*}{\columnwidth}{@{\extracolsep{\fill}} llccc}
    \toprule[1pt]
        Method                                             & Metric                & Replica       & ScanNet++         & Indoor \\ \hline
        \multirow{5}{*} 
        {\makecell{GauS-SLAM}}      & FPS$\uparrow$           & 0.97              & \ding{53}         & 0.83  \\
                                                           & Memory$\downarrow$      & 13682             & \ding{53}         &  14930   \\ 
                                                           & PSNR$\uparrow$          & \textbf{39.96}     &  \ding{53}   & 29.18    \\
                                                           & SSIM$\uparrow$          & \textbf{0.980}      &  \ding{53}   & 0.881    \\
                                                           & LPIPS$\downarrow$       & \textbf{0.052}     &  \ding{53}   & 0.277    \\ \hdashline
        \multirow{5}{*} 
        {\makecell{RTG-SLAM}} & FPS$\uparrow$           & 17.31              &  \underline{4.58} & 14.82    \\
                                                           & Memory$\downarrow$      & \textbf{2882}     &  \textbf{5120}    & \textbf{3319}    \\ 
                                                           & PSNR$\uparrow$     & 34.73     &  \textbf{29.08}       & 30.03    \\
                                                           & SSIM$\uparrow$     & \underline{0.972}     &  \underline{0.911}       & \underline{0.908}    \\
                                                           & LPIPS$\downarrow$    & 0.119     &  \underline{0.209}       & 0.264    \\ \hdashline
        \multirow{5}{*} 
        {\makecell{GSFusion}}        
                                                            & FPS$\uparrow$          & 14.16          & 4.26           & 15.33   \\                
                                                           & Memory$\downarrow$     & 7977          & \underline{6415}  & 7817   \\ 
                                                           & PSNR$\uparrow$         & 35.12         &  28.84            & \textbf{30.99}    \\
                                                           & SSIM$\uparrow$         & 0.953         &  0.897            & \textbf{0.919}    \\
                                                           & LPIPS$\downarrow$      & 0.148         &  0.216            & 0.238    \\ \hdashline
        \multirow{5}{*} 
        {\makecell{GS-ICP SLAM}}  & FPS$\uparrow$           & \underline{166.39} & \ding{53}         & \underline{114.18}    \\
                                                           & Memory$\downarrow$      & 5893              & \ding{53}         & 10251   \\ 
                                                           & PSNR$\uparrow$          & 33.15     &  \ding{53}   & 27.05    \\
                                                            & SSIM$\uparrow$         & 0.934     &  \ding{53}   & 0.874    \\
                                                            & LPIPS$\downarrow$      & 0.158     &  \ding{53}   & 0.287    \\ \hdashline
        \multirow{5}{*} 
        {\makecell{GPS-SLAM\\(ours)}}                                  & FPS$\uparrow$           & \textbf{252.64}   &  \textbf{79.18}   & \textbf{151.00}    \\

                                                           & Memory$\downarrow$      & \underline{3979}  &  8870             & \underline{4098}   \\
                                                           & PSNR$\uparrow$     & \underline{37.24}     &  \underline{28.97}       & \underline{30.19}   \\
                                                            & SSIM$\uparrow$     & 0.960     &  \textbf{0.918}       & 0.907   \\
                                                            & LPIPS$\downarrow$    & \underline{0.103}     &  \textbf{0.206}       & \textbf{0.230}   \\ 
                                                           
    \bottomrule[1pt]
    \end{tabular*}
\end{table} 

Furthermore, we conducted a detailed time cost analysis on Replica \emph{office0}. We report in Table~\ref{tab:time_analysis} the mapping time per frame and the optimization time per frame. We also record the optimization time per iteration, the total iteration count, the total Gaussian count, the PSNR, and the overall system fps. The slow reconstruction of GauS-SLAM and RTG-SLAM arises from their high iteration counts and long optimization times per iteration. GS-ICP SLAM exhibits the fewest iteration counts; however, this is because its tracker operates so quickly that its mapper does not fully optimize the Gaussians, and the rendering quality is significantly degraded. GSFusion accelerates the optimization time by reducing the number of Gaussians. However, its Gaussian initialization mechanism, which relies on a quadtree scheme for each input RGB image, is excessively time-consuming (53.5 ms per frame). In contrast to GSFusion, our method eliminates the complex Gaussian initialization process and achieves superior reconstruction results with fewer iterations.

\begin{table*}[t!]
    \centering
    \caption{Time costs on Replica \emph{office0} (2000 frames). * indicates that we manually remove GauS-SLAM's tracking time when calculating the overall FPS for a fair comparison, even though mapping and tracking are implemented in a single thread in its published code.} 
    \label{tab:time_analysis}
   \begin{tabular}{lrrrrrrr}
    \toprule[1pt]
        Method                                & \makecell{Mapping\\/Frame}  &\makecell{Optimization\\/Frame} &\makecell{Optimization\\/Iteration} &\makecell{Iteration\\Count}   &\makecell{Gaussian\\Count} &\makecell{PSNR$\uparrow$} &FPS$\uparrow$ \\ \hline
        GauS-SLAM       & 341.2 ms   & 336.9 ms  & 14.3 ms & 47120 & 901685 & \textbf{43.11} & 2.93*       \\ 
        RTG-SLAM   & 58.3 ms     & 37.6 ms   & 4.5 ms & 16700  & 268779 & 38.62 & 17.15          \\
        GSFusion          & 65.0 ms     & 6.0 ms  & \underline{1.2 ms} &  10000 & \textbf{112282} & 37.64 & 15.37          \\
        GS-ICP SLAM    & \underline{5.7 ms}      & \underline{3.6 ms}  & 8.5 ms & \textbf{843}   & 1679211& 37.33 & \underline{174.20} \\ 
        GPS-SLAM (ours)                                  & \textbf{2.6 ms}      & \textbf{2.2 ms}   & \textbf{1.1 ms} & \underline{4000}   & \underline{137200} & \underline{41.15} & \textbf{380.72}          \\
    \bottomrule[1pt]
    \end{tabular}
 \end{table*}

\subsubsection{Tracking accuracy} We report the average tracking accuracy on both the synthetic Replica Dataset and the real-world TUM-RGBD Dataset in Table~\ref{track_acc_TUM}. Here, we also include three classic traditional SLAM systems: Kintinuous~\cite{Kintinuous}, BundleFusion~\cite{bundlefusion}, and ORB-SLAM~\cite{ORBSLAM2} for a more comprehensive comparison. GauS-SLAM demonstrates high accuracy on both datasets. However, its tracker is slow because it optimizes the camera pose by minimizing the difference between the Gaussian-rendered color and the input image. Among the remaining methods, our method achieves comparable results on the Replica Dataset. However, on the TUM-RGBD Dataset, which contains inaccurate depth input, our method exhibits relatively lower accuracy, like all ICP-based methods.
\begin{table}[t!]
    \centering
    \caption{Comparison of tracking accuracy on the TUM-RGBD and Replica Datasets. Numbers represents absolute trajectory error (ATE) root mean square error (RMSE) in cm.}
    \label{track_acc_TUM}
    \begin{tabular}{l l r r r}
    \toprule[1pt]
        Category                                &Method                                  & Replica         & TUM   \\ \hline
        \makecell{Image Loss}                   &GauS-SLAM          & \textbf{0.06}            & 1.54   \\ \hdashline
        \multirow{3}{*}
        {\makecell{ICP  Loss}}                  &GS-ICP SLAM      & \underline{0.15}            & 2.89   \\
                                                &Kintinuous        & 0.29            & 3.20   \\
                                                &GPS-SLAM (ours)                                    & 0.17            & 3.08  \\ \hdashline
        \multirow{3}{*}
        {\makecell{ICP Loss \&\\Backend}}       &BundleFusion    & 0.16            & 2.07   \\
                                                &ORB-SLAM2           & 0.68            & \textbf{1.00}   \\
                                                &RTG-SLAM     & 0.18            & \underline{1.06}   \\
    \bottomrule[1pt]
    \end{tabular}
\end{table}

\subsubsection{Rendering quality} 
\begin{figure*}[t!]
    \centering
    \includegraphics[width= \textwidth]{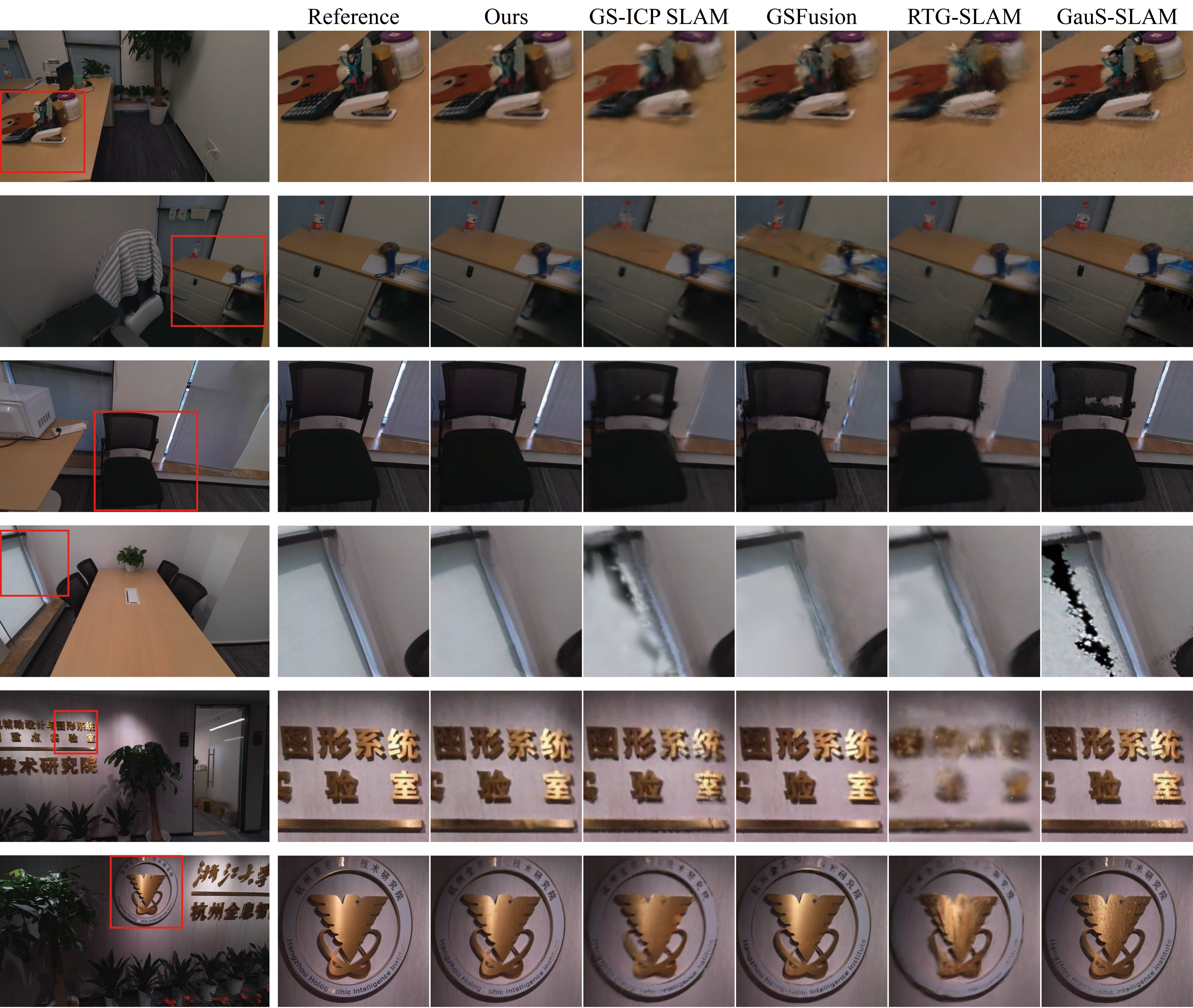}
    \caption{Qualitative comparison of rendering quality on our indoor dataset. Our method achieves high-fidelity reconstruction, while other baselines produce blurred results or significant artifacts. Please note that the poor reconstruction quality of GS-ICP SLAM results from its mapping operation being slower than its tracking.}
    \label{fig:Indoor_compare}
\end{figure*}

We compared rendering quality on the Replica, ScanNet++, and Indoor Datasets; results are shown in Table~\ref{tab:time_and_memory_and_render}. Since the ScanNet++ dataset contains many areas with missing depth, and consistent with previous methods \cite{GS-ICP_SLAM, GauS-SLAM}, we only calculate image metrics within regions where depth is available. Our method achieves results comparable to state-of-the-art approaches. Moreover, it has significantly improved image quality compared to GS-ICP SLAM, which is the only other SLAM system running at over 100 fps. The results of GS-ICP SLAM are worse than those reported in its original paper. This is because its tracker runs faster in our environment, resulting in fewer Gaussian optimization iterations. Qualitative comparisons on our indoor dataset are shown in Fig.~\ref{fig:Indoor_compare}. All other methods exhibit noticeable artifacts and blurring in some regions. Our Gaussian-plus-SDF representation overcomes this limitation, achieving high-quality reconstruction in these regions. We also evaluated the novel view synthesis quality of our method using the test split of ScanNet++. The results are presented in Table~\ref{test_rendering_scannet++} and demonstrate that our method can also achieve comparable results in novel view synthesis.

\begin{table}
    \centering
    \caption{Comparison of novel view synthesis on the ScanNet++ Dataset.} 
    \label{test_rendering_scannet++}
    \begin{tabular}{l r r r }
    \toprule[1pt]
        Method                              & PSNR$\uparrow$      & SSIM$\uparrow$       & LPIPS$\downarrow$      \\ \hline
        GSFusion        & 24.73     & \textbf{0.884}      &   \textbf{0.264}         \\ 
        RTG-SLAM & \underline{25.07}     & 0.852      & 0.286           \\ 
        GPS-SLAM (ours)                                & \textbf{25.81}     & \underline{0.872}      & \underline{0.278}          \\
    \bottomrule[1pt]
    \end{tabular}
\end{table}

\begin{figure}[t!]
    \centering
\includegraphics[width=\columnwidth]{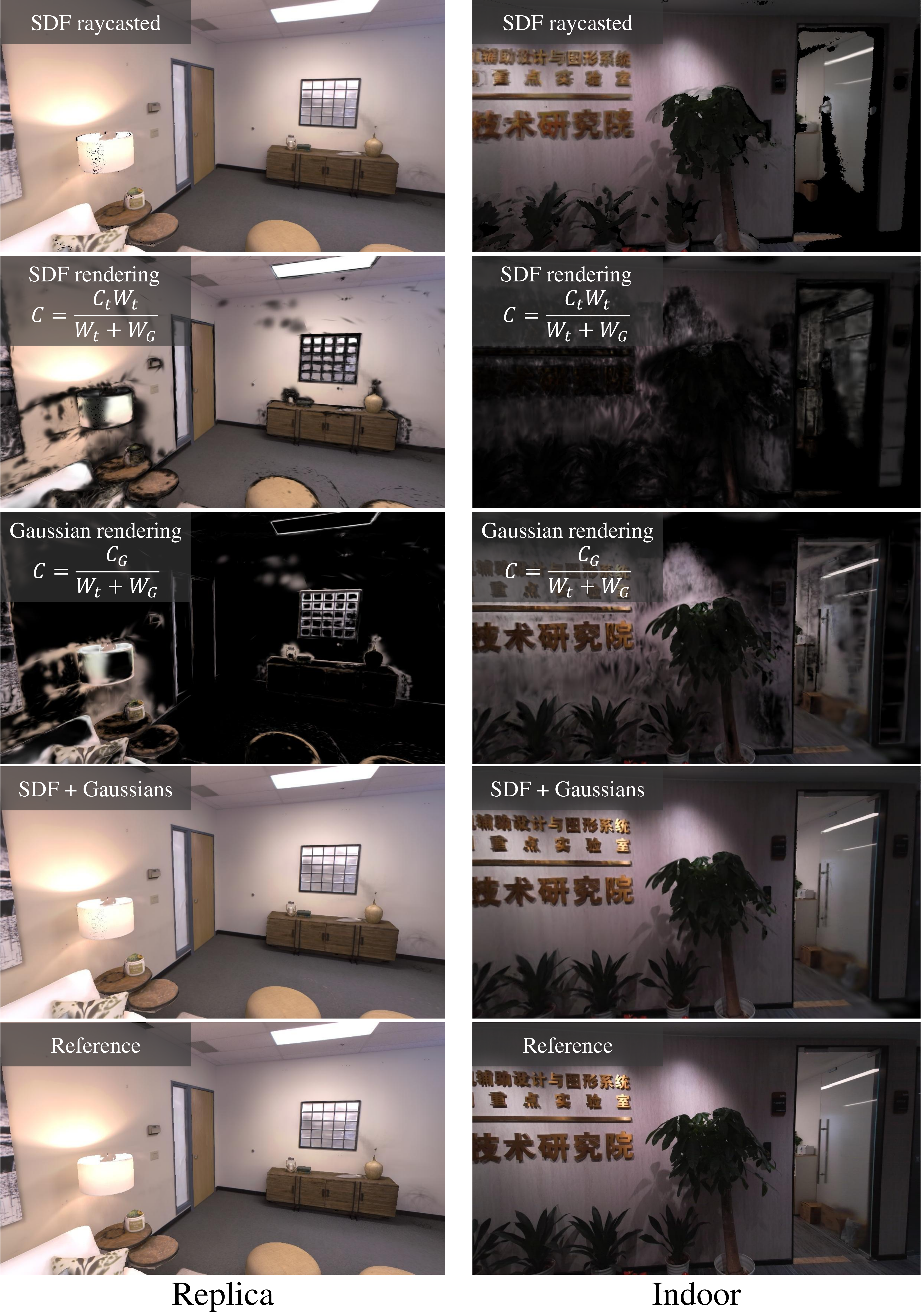}
    \caption{Comparison including images rendered with only SDF or only 3D Gaussians.}
    \label{fig::render_method_ablation}
\end{figure}

\subsubsection{Geometry quality} Following NICE-SLAM, we use the following metrics to evaluate the scene geometry on ScanNet++: \emph{Accuracy}, \emph{Completion}, \emph{Accuracy Ratio} [\textless3~cm], and \emph{Completion Ratio} [\textless3~cm]. Following \cite{nice_slam}, we remove unseen regions that are not inside any camera's frustum. We extract meshes from the reconstructed SDF volume using marching cubes for GSFusion and our method. For RTG-SLAM, we uniformly sample an equal number of points from the reconstructed Gaussians for evaluation. To eliminate the impact of tracking accuracy, we use the ground truth camera pose for reconstruction. Results are reported in Table~\ref{geometry_accuracy_scannet++}. Our method outperforms GSFusion because its default voxel size is 1~cm, while our voxel size is 0.5~cm. This demonstrates that good geometry can be reconstructed using high-precision SDF fusion alone, without the need for a Gaussian representation.

\begin{table}[t]
    \centering
     \caption{Comparison of geometric accuracy on the ScanNet++ Dataset.} 
    \label{geometry_accuracy_scannet++}
    \begin{tabular}{l r r r}
    \toprule[1pt]
        Metric              & GSFusion           & RTG-SLAM               & GPS-SLAM        \\ \hline
        Accuracy $\downarrow$     & 1.38               & \underline{0.94}       & \textbf{0.60}          \\
        Accuracy Ratio$\uparrow$ & 92.51              & \underline{96.87}      & \textbf{99.93}         \\
        Completion $\downarrow$    & 1.26               & \underline{1.09}       & \textbf{0.61}          \\
        Completion Ratio$\uparrow$& 95.74              & \underline{97.31}      & \textbf{99.50}         \\
    \bottomrule[1pt]
    \end{tabular}
\end{table}

\subsection{Ablation and Related Studies}
\subsubsection{Rendering using only SDF or Gaussians} 
As  Fig.~\ref{fig::render_method_ablation} shows, using the synthetic Replica Dataset, SDF fusion achieves good quality. In this case, our method requires only a small number of Gaussians to fit some high-frequency details. However, on our real-world indoor dataset, SDF fusion introduces significant artifacts and incompleteness due to more view-dependent appearances and missing depths. Consequently, our method uses more Gaussians to represent this information, resulting in slower performance on real-world datasets.

\begin{table}[t!]
    \centering
    \caption{Quantitative comparisons of different SDF voxel sizes using Replica \emph{room0}. Geometric quality is evaluated using the chamfer distance (CD) metric.} 
    \label{tab:voxel_size_ablation_studies}
    \begin{tabular}{l c c c}
    \toprule[1pt]
        Metric              & 0.5 cm                & 1 cm                   & 2 cm           \\ \hline
        PSNR$\uparrow$      & \textbf{34.73}       & \underline{34.09}     & 33.01         \\
        FPS$\uparrow$       & \textbf{133.88}      & \underline{107.48}    & 96.02         \\
        ATE$\downarrow$     & \textbf{0.16 cm}      & \underline{0.21 cm}    & 0.26 cm        \\
        CD$\downarrow$      & \textbf{1.72 cm}      & \underline{1.77 cm}    & 1.97 cm        \\
        Gaussian Size       & \textbf{155 MB}       & \underline{193 MB}     & 222 MB         \\
        Mesh Size           & 2284 MB               & \underline{560 MB}     & \textbf{138 MB}\\
    \bottomrule[1pt]
    \end{tabular}
\end{table}

\subsubsection{SDF voxel size} 
\begin{figure}[t!]
    \centering
\includegraphics[width=\columnwidth]{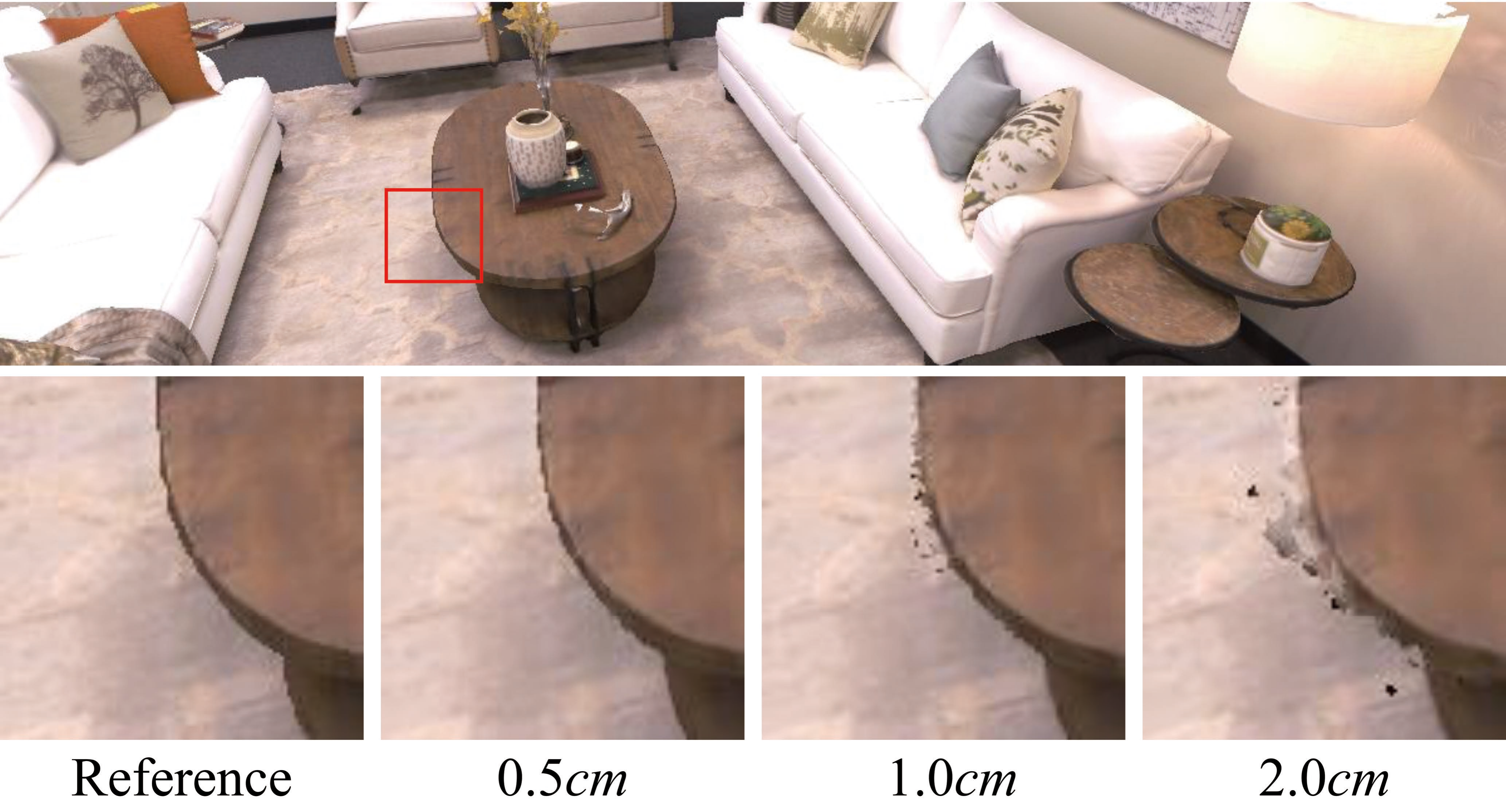}
    \caption{Effect of SDF voxel size on Replica \emph{room0}.}
    \label{fig::voxel_size_ablation_studies}
\end{figure}

In our framework, the SDF voxel size plays a crucial role, influencing not only the quality of the reconstructed geometry but also the Gaussian optimization and model size. We performed experiments with different voxel sizes using Replica \emph{room0}. The results are summarized in Table~\ref{tab:voxel_size_ablation_studies} and illustrated in Fig.~\ref{fig::voxel_size_ablation_studies}. The experiments demonstrate that smaller voxel sizes result in higher tracking accuracy and improved image quality, especially in border areas. Moreover,  reconstruction speed is not sacrificed. This is because the primary bottleneck of our system is Gaussian optimization, and more accurate SDF reconstruction reduces the number of Gaussians needed, thus accelerating the optimization process.

\begin{figure}[t!]
    \centering
\includegraphics[width=\columnwidth]{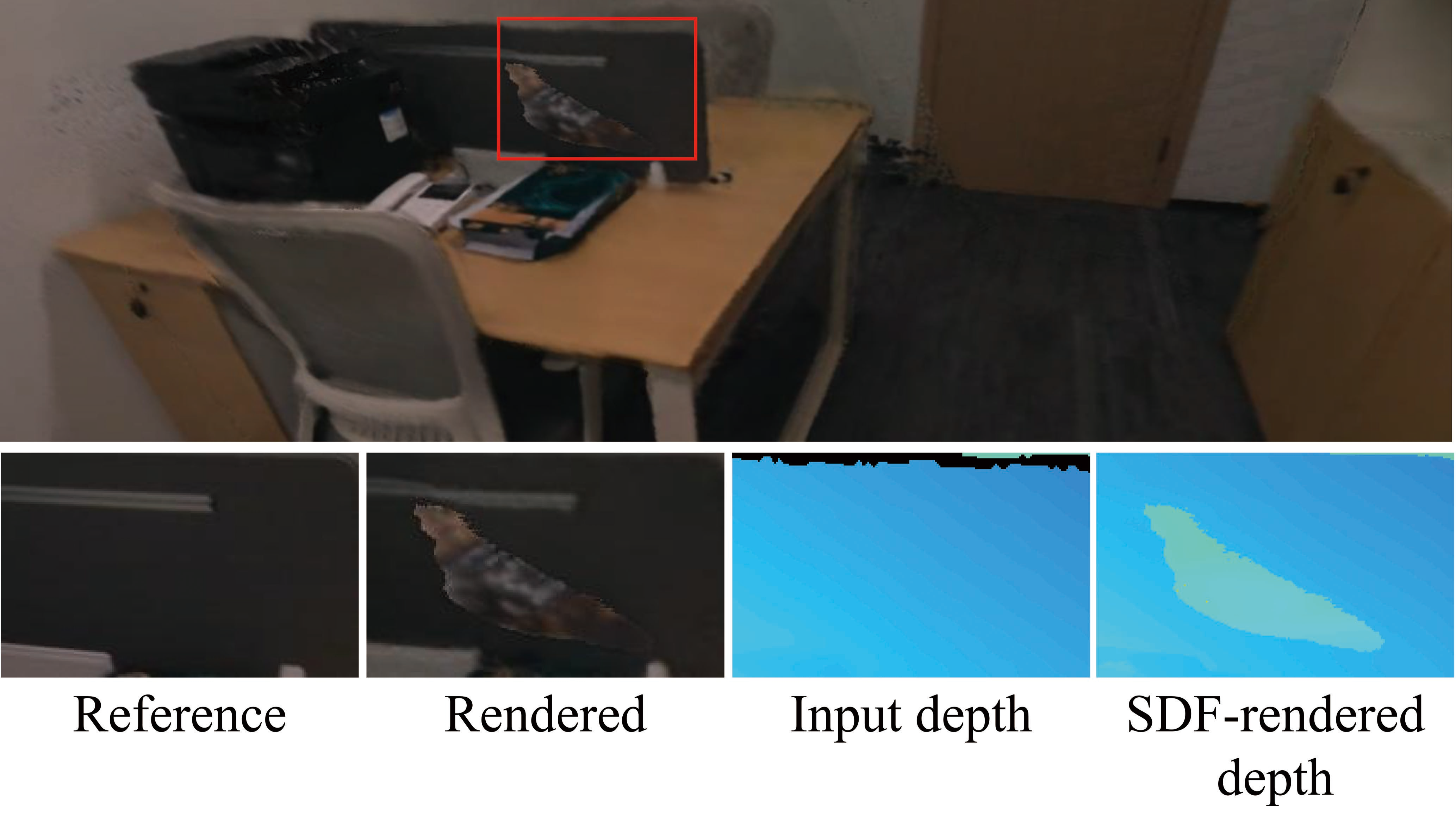}
    \caption{\peng{A failure. The thin panels exhibit holes during SDF reconstruction, which causes colors to be incorrectly rendered onto the opposite side.}}
    \label{fig::geo_error_limitation}
\end{figure}

\subsubsection{Sorting-free Gaussian rendering} Our sorting-free Gaussian rendering also contributes to the ultra-fast performance. We test the runtime performance of two rendering methods on the \emph{activity room} scene of our indoor dataset, reporting the forward and backward times in Table~\ref{tab:sorting_free_ablation_studies}. All results were averaged over three trials to ensure statistical reliability. Sorting-free Gaussian rendering accelerates the forward pass by 10\% and the backward pass by 19\%, resulting in an overall system speedup of 17\%.

\begin{table}[t!]
    \centering
    \caption{Benefits of sorting-free Gaussian rendering on our indoor \emph{activity room}. Our sorting-free Gaussian rendering simultaneously speeds up both forward and backward process.}
    \label{tab:sorting_free_ablation_studies}
   \begin{tabular}{c c c c c c c c c}
    \toprule[1pt]
        Rendering method   & Forward            & Backward     & System FPS     \\ \hline
        With sort       & 0.97 ms             & 1.87 ms       & 152.86 \\ 
        Without sort    & \textbf{0.87 ms}             & \textbf{1.51 ms}       & \textbf{179.40} \\ 
    \bottomrule[1pt]
    \end{tabular}
 \end{table} 

\section{Conclusions}
We have presented a real-time 3D reconstruction system which  enables  ultra-fast, high-fidelity reconstruction. The core of our system is a novel representation called Gaussian-plus-SDF, which can minimize the computational overhead of Gaussian optimization. We have designed highly efficient Gaussian insertion, Gaussian optimization and Gaussian removal based on Gaussian-plus-SDF in the SLAM system. Our system achieves an order-of-magnitude faster reconstruction speed than state-of-the-art methods, with comparable reconstruction quality. 

Our system currently only includes front-end camera tracking, which can lead to drift in large-scale scenes. Incorporating global pose optimization like Loopy-SLAM might address this issue. Additionally, our Gaussian rendering relies on depth culling provided by SDF. \peng{As  Fig.~\ref{fig::geo_error_limitation}} shows, significant geometric errors in the SDF reconstruction can degrade rendering quality.
In  future,  we plan to explore the ultra-fast reconstruction of large-scale outdoor scenes with LiDAR sensors. We believe Gaussian-plus-SDF holds the promise of being compatible with LiDAR data.

\appendix

\numberwithin{equation}{section}
\subsection*{Appendix}
\section{Gaussian Initialization}\label{app:initialization_details}
Here we explain how we compute the scales $\mathbf{s}_\mathbf{u}$ of each newly added Gaussian for pixel $\mathbf{u}$ in detail. Each new Gaussian is initialized as a thin circular disc as in RTG-SLAM. However, RTG-SLAM calculates the scales using all Gaussians in the scene, resulting in an excessively large $k$-NN scope and impacting speed. Therefore, we only use the selected pixels in  $\mathbf{V}_k^{*,g}$. Specifically, we select the three vertices $\mathbf{V}_k^{*,g}(\mathbf{u}_1), \mathbf{V}_k^{*,g}(\mathbf{u}_2), \mathbf{V}_k^{*,g}(\mathbf{u}_3)$  closest to the pixel $\mathbf{V}_k^{*,g}(\mathbf{u})$ , where $\mathbf{u}_1,\mathbf{u}_2,\mathbf{u}_3\in M{\widetilde{\mathbf{{u}}}}$. The scales of $\mathcal{G}_u$ are initialized based on the following formula
\begin{align}
       &\mathbf{s}_{\mathbf{u},1}= 
       \begin{cases}
       0.1\quad \text{if}\; \sqrt{\frac{1}{3}\sum_{i=1}^3 \left(||\mathbf{V}_k^{*,g}(\mathbf{u}) - \mathbf{V}_k^{*,g}(\mathbf{u}_i)||\right)}\geqslant 0.1 \\
       \sqrt{\frac{1}{3}\sum_{i=1}^3 \left(||\mathbf{V}_k^{*,g}(\mathbf{u}) - \mathbf{V}_k^{*,g}(\mathbf{u}_i)||\right)} \quad \text{otherwise}, \nonumber
       \end{cases}\\
       &\mathbf{s}_{\mathbf{u},2}= \mathbf{s}_{\mathbf{u},1}, \quad \mathbf{s}_{\mathbf{u},3} = 0.1\mathbf{s}_{\mathbf{u},1}.
\end{align}
Here, we truncate the maximum scale to below 0.1 to avoid large-scale outliers.

\section{Gaussian Optimization}\label{app:optimization_details}
Although the sort-free rasterization method allows us to launch threads per Gaussian, the significant variation in  Gaussian scale means that directly launching a thread for each Gaussian would lead to substantial computational imbalance between threads. To address this, we calculate the number of pixels covered by each Gaussian based on its radius in pixel space. We then divide the Gaussians into groups of a fixed size. During backpropagation, threads are launched per group, ensuring a consistent number of iterations per thread and maximizing parallel efficiency.

\section{Implementation Details}\label{app:implement_details}
Now we list all the parameters in detail. For Gaussian optimization learning rates, we set $\mathrm{lr}_\mathrm{position}=0.00016$, $\mathrm{lr}_\mathrm{SH0}=0.0025$, $\mathrm{lr}_\mathrm{\alpha}=0.05$, $\mathrm{lr}_\mathrm{scale}=0.005$ and $\mathrm{lr}_\mathrm{rotation}=0.001$ for all datasets. The learning rate for other SH coefficients is 0.0005. The motion thresholds for keyframe creation, $\delta_\mathrm{angle}$ and $\delta_\mathrm{move}$, are set to $30\degree$ and 0.3 m, respectively.

\section{Dataset Details}\label{app:dataset_details}
We used a laptop with an Intel i7 10870-H CPU and nVidia 3070 Laptop GPU connected to an Azure Kinect RGB-D camera for data acquisition. The RGB-D images captured by the camera were transmitted to a desktop computer through a wireless network, and the desktop computer completed the SLAM computations. Results were sent back to a viewer on the laptop for visualization. A statistical summary of our indoor dataset is given in Table \ref{tab:statistics_indoor_dataset}. 

\begin{table}[t]
    \caption{Statistical summary of our indoor dataset} 
    \centering
    \setlength{\tabcolsep}{2pt} 
    \begin{tabular}{l c c c c c}
    \toprule[1pt]
        Statistic       & signboard & activity room & office0 & office1 & office2 \\ \hline
        Trajectory (m) & 13.8      & 13.3          & 21.7    & 15.2    & 18.2    \\
        Area (m$^2$)       & 37.3      & 30.3          & 88.5    & 32.5    & 39.8    \\
        Frames            & 2200      & 2680          & 2950    & 2150    & 2500    \\
    \bottomrule[1pt]
    \end{tabular}
    \label{tab:statistics_indoor_dataset}
\end{table} 

\subsection*{Availability of data and materials}
We used the publicly available Replica, TUM-RGBD and ScanNet++ Datasets in our experiments. Our own indoor dataset can be downloaded from \url{https://gapszju.github.io/GPS-SLAM.}

\subsection*{Funding}
This research was supported by NSF China (U23A20311, 62421003).

\subsection*{Author contributions}
Kun Zhou introduced the idea, and designed the algorithm with Tianjia Shao. Zhexi Peng implemented the algorithm, designed and conducted the experiments, and prepared the draft paper. Kun Zhou and Tianjia Shao had regular discussions with Zhexi Peng, provided guidance and suggestions, and revised the paper.

\subsection*{Acknowledgements}
The authors appreciate the support from Adobe Research and the XPLORER Prize.

\subsection*{Declaration of competing interest}
The authors have no competing interests to declare that are relevant to the
content of this article. Kun Zhou is an Associate Editor of this journal.

\subsection*{Electronic supplementary material}
Electronic supplementary material is available in the online version of this article.

\bibliographystyle{CVMbib}
\bibliography{refs}

\subsection*{Author biography}
\begin{biography}[pzx]{Zhexi Peng} received his B.S. degree in information and computing sciences from Beijing University of Posts and Telecommunications. He is currently a Ph.D. student at Zhejiang University. His research interests cover computer vision, SLAM, and machine learning.
\end{biography}
\vspace*{2.6em}

\begin{biography}[kunzhou]{Kun Zhou} is a Cheung Kong Professor of Computer Science at Zhejiang University and the Director of the State Key Lab of CAD\&CG. He received his Ph.D. degree from Zhejiang University, then spent six years with Microsoft Research Asia, serving as a lead researcher of the graphics group before returning to Zhejiang University. He was named one of the world's top 35 young innovators by MIT Technology Review (2011), and received an Asiagraphics Outstanding Technical Contributions Award (2022) and an ACM SIGGRAPH Test-of-Time Award (2024). He is a Fellow of IEEE and ACM.
\end{biography}
\vspace*{2.6em}

\begin{biography}[tianjia_crop]{Tianjia Shao} received his B.S. from the Department of Automation, and his Ph.D. in computer science from the Institute for Advanced Study, both in Tsinghua University. He is currently a ZJU100 Young Professor in the State Key Laboratory of CAD\&CG, Zhejiang University. Previously he was a Lecturer (= Assistant Professor) in the School of Computing, University of Leeds, UK. His current research focuses on 3D scene reconstruction, digital human creation, and 3D AIGC.
\end{biography}

\end{document}